\documentclass[letterpaper, 10 pt, conference]{ieeeconf}  

\IEEEoverridecommandlockouts                              

\let\labelindent\relax

\usepackage[
    style=ieee,
    doi=false,
    isbn=false,
    url=false,
    eprint=false,
    backend=bibtex,
    natbib=true,
    minnames=1,
    maxcitenames=1
    ]{biblatex}

\usepackage[shortcuts]{extdash}
\usepackage{graphicx}
\usepackage{mathtools}
\usepackage{dcolumn}
\usepackage{gensymb}
\usepackage{perl_acronyms}
\usepackage{xcolor}
\usepackage{amsfonts}
\usepackage{booktabs}
\usepackage{siunitx}
\usepackage[inline]{enumitem}
\usepackage{array}
\usepackage{flushend}
\newcolumntype{P}[1]{>{\centering\arraybackslash}p{#1}}

\title{\LARGE \bf
SilhoNet-Fisheye: Adaptation of A ROI Based Object Pose Estimation Network to Monocular Fisheye Images
}

\author{Gideon Billings$^{1}$ and Matthew Johnson-Roberson$^{1}$
\thanks{$^{1}$Gideon Billings and Matthew Johnson-Roberson are with DROP lab, Department of Naval Architecture and Marine Engineering,
        University of Michigan, 2600 Draper Dr. Ann Arbor, MI 48109, USA
        {\tt\small gidobot@umich.edu, mattjr@umich.edu}}%
}

\begin{document}

\maketitle

\begin{abstract}
\label{sec:abstract}

There has been much recent interest in deep learning methods for monocular image based object pose estimation. While object pose estimation is an important problem for autonomous robot interaction with the physical world, and the application space for monocular-based methods is expansive, there has been little work on applying these methods with fisheye imaging systems. Also, little exists in the way of annotated fisheye image datasets on which these methods can be developed and tested. The research  landscape is even more sparse for object detection methods applied in the underwater domain, fisheye image based or otherwise. In this work, we present a novel framework for adapting a ROI-based 6D object pose estimation method to work on full fisheye images. The method incorporates the gnomic projection of regions of interest from an intermediate spherical image representation to correct for the fisheye distortions. Further, we contribute a fisheye image dataset, called UWHandles, collected in natural underwater environments, with 6D object pose and 2D bounding box annotations.

\end{abstract}

\section{INTRODUCTION}
\label{sec:introduction}

The advantages of fisheye imaging systems in robotics applications has long been recognized. With technological improvements in imaging sensor resolution and dynamic range, fisheye cameras can capture significantly greater information about the surrounding environment without appreciably increasing the imaging sensor footprint, compared to their perspective model counterparts. However, little work has been done on applying CNN based methods to the problem of 6D object pose estimation on full fisheye images. Dealing with fisheye images is challenging, due to the large distortions and viewpoint ambiguities arising from the wide field of view. We address the problem of 6D object pose estimation in full fisheye images by proposing a method whereby the image is first projected to the surface of a sphere, where we mathematically define a consistent apparent viewpoint which the network is trained to predict. The true orientation relative to the fisheye frame can then be recovered using the predicted translation. The gnomonic projection is used in our method to undistort the \ac{ROI} from the sphere surface, and we investigate applying this projection both before and after the feature extraction stage.

Further, while deep learning has greatly advanced the state of-the-art in terrestrial based visual methods for object detection and pose estimation, the challenges of collecting underwater datasets for training these methods and the limited access to underwater environments has hindered progress in leveraging these advancements for underwater applications. We address this challenge by introducing an easy and efficient method for collecting fiducially grounded monocular images of underwater scenes containing known objects, and we contribute a tool for annotating the 6D object poses and bounding boxes in the image sequences. We also contribute an underwater visual dataset of three different graspable handles with annotated ground truth poses and bounding boxes. The dataset was collected with a fisheye camera mounted on the wrist of an ROV manipulator, with objects appearing in different arrangements in diverse natural seafloor environments.

In summery, we present the following contributions:
\begin{enumerate*}
    \item A framework for adapting \ac{ROI}-based networks for predicting 6D object pose from monocular images to work on full fisheye images, through an intermediate mapping onto a sphere and \ac{ROI} processing through the gnomonic projection. This adaptation is demonstrated with the SilhoNet method presented in ~\cite{billings2019silhonet};
    \item UWHandles, an underwater monocular fisheye image dataset of handle objects, with annotated 6D pose and 2D bounding boxes, collected in natural seafloor environments. We also release the annotation tool, VisPose, used to process the dataset;
\end{enumerate*}

The rest of this paper is organized in the following sections: Section II discusses related work; Section III presents our methods for adapting SilhoNet to the fisheye domain; Section IV presents the UWHandles dataset; Section V presents the experimental results; and Section VI concludes the paper.


\section{RELATED WORK}
\label{sec:related}
In general, state-of-the-art works that apply CNN methods to full fisheye images process the raw images directly through the network without special consideration of the fisheye distortions~\cite{7995725, 8906325, cnn-based-fisheye}. These networks are mostly applied to the problems of segmentation or \ac{ROI} detection in the fisheye images. Due to the sparsity of available benchmarking datasets for fisheye images, these works report their results on synthetic datasets, generated by projecting perspective images to distorted fisheye images. \citet{zhu-fisheye} used a CNN in the prediction of ground vehicle positions relative to an aerial fisheye imaging platform. They directly train the CNN on the raw fisheye images to generate \ac{ROI} proposals. They assume the detected object is on the ground plane and fuse measurements from height and orientation sensors on the camera platform to recover only the object's 3D translation in the world. \citet{7905998} extended the Cascaded Pose Regression algorithm to estimate the 3D pose of mice in fisheye images from detected 3D keypoints. However, their method incorporates priors about the structured lab environment, and the fisheye camera is fixed in the scene, allowing them to easily segment the mice from the background image. In contrast to these works, our method incorporates knowledge of the fisheye distortion model through a spherical mapping, which improves network performance and is also necessary to create visually consistent pose annotations which can be regressed directly from \ac{ROI} proposals across the full fisheye field of view. Further, we report the performance of our method on a real fisheye dataset captured in a natural unstructured environment.

Closely related to fisheye image processing is the extensive body of work on omni-directional imaging, as both fisheye and omni-directional image distortions can be represented on a sphere. Beyond naively applying CNNs directly to a flattened equirectangular projection of an omni-directional image, which has been shown to suffer from the nonlinear distortions of the spherical mapping to the plane and attain sub-optimal performance~\cite{shan2019discrete}, the methods of dealing with omni-directional distortions can be roughly categorized under three approaches: generating multiple perspective projections from the sphere, such as cube map, and processing each projection separately through the CNN~\cite{monroy2018salnet360}; adapting the kernel sampling locations based on a spherical distortion model or a learned mapping~\cite{zhao2018distortion, coors2018spherenet, su2017learning, su2019kernel}; re-sampling the spherical image based on a uniform sampling geometry such as the icosahedron, and processing the spherical representation with specialized convolution operations~\cite{jiang2019spherical, eder2019convolutions, lee2019spherephd, zhao2019spherical}; or transforming the spherical feature signals and convolution operations into the spectral domain, typically by representation of the spherical image as a graph~\cite{perraudin2019deepsphere, khasanova2017graph, cohen2018spherical}. Methods that operate on multiple perspective projections suffer from discontinuities at the projection borders, due to variance in feature appearance on different tangent plane mappings. Methods that operate on graphical representations of the sphere in the spectral domain are memory limited in scaling to full resolution images and have some level of rotation invariance in the convolution response function, which is undesirable when regressing 6D object pose. Methods that re-sample the convolution kernel sampling location based on a learned or distortion based mapping are most relevant to our work. The methods of \citet{coors2018spherenet} and \citet{zhao2018distortion} sample regular kernel locations on a tangent plane and then project the sampling locations to the spherical surface, encoding the spherical distortions directly into the convolution operation. \citet{zhao2019reprojection} adapts a region proposal network with the distortion aware convolutions of \cite{coors2018spherenet, zhao2018distortion} in a two-stage architecture to predict region proposals from omni-directional images. However, these distortion aware convolutions are designed to operate on full 360\degree images. Because fisheye images represent only a partial view of the sphere, they can also be analyzed under different planar projections than omni-directional images. Further, application of omni-directional CNNs to 6D object pose estimation is so far lacking in the literature.

The body of work applying CNN methods to underwater imagery is mostly limited to the problems of species detection and classification~\cite{feng2019underwater, chen2019underwater, rathi2017underwater, lopez2020video, konovalov2019underwater, xu2018underwater, marini2018tracking, salman2019automatic, moniruzzaman2017deep}, or underwater image correction~\cite{li2017watergan, li2018emerging} on perspective images. \citet{kuang2019depth} used a simple color distortion model based on image depth to generate a synthetic dataset of omni-directional images that were color cast as though captured underwater. They trained a distortion aware CNN to predict image depth from an omni-directional image, and reported results on their synthetic dataset. While they did not test with real omni-directional data, the perspective image equivalent of their method performed very poorly on real underwater images. Most related to our work in the underwater domain is the work of~\citet{jeon2019underwater}, who proposed a CNN based method for underwater object detection and pose estimation, using a synthetic dataset generated from CAD models to train the network. However, the objects used in their dataset were very simple, and their tests were limited to tank environment with high contrast between the object models and the scene background. Further, they only regressed the 3D orientation of the detected objects. Also related to our work is \citet{nielsen2019evaluation}, where a PoseNet CNN was trained to regress the 6D pose of a mock-up sub-sea connector relative to a small ROV. The dataset was collected in a tank environment, with high contrast between the connector target and the low featured background. In contrast to these works, our method addresses the problem of full 6D object pose estimation from monocular underwater images captured in wild unstructured environments. Further, our method is applied to full view fisheye images, which capture a significantly greater field of view over perspective images. Also, our dataset is composed with visually challenging handle objects used to manipulate ROV tools in real life applications.

\begin{figure*}[t!]
    \centering
    \includegraphics[width=\textwidth]{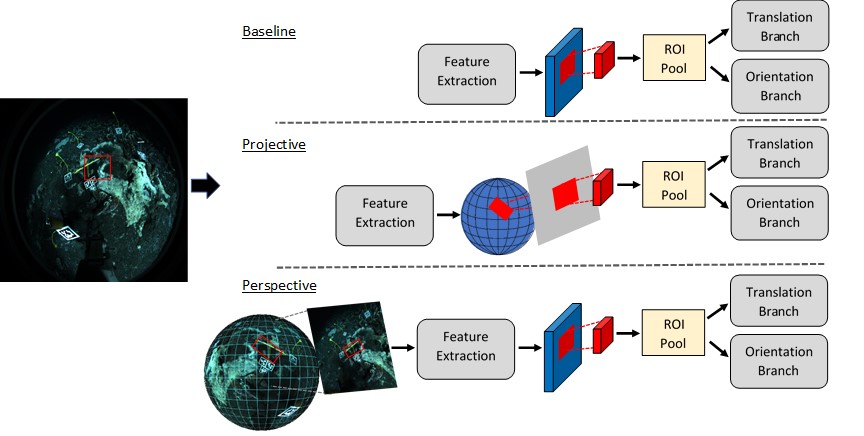}
            \vspace{-20pt}
    \caption{Overview of the three different SilhoNet~\cite{billings2019silhonet} adaptations for processing full fisheye images. The Baesline method processes the raw fisheye image directly through the unmodified network. The Projective variant processes the raw fisheye image through the feature extraction stage and then projects the features within the \ac{ROI} through a spherical mapping to the tangent plane centered on the \ac{ROI}, before processing the features through the ROI-pooling stage. The Perspective adaptation maps the fisheye image to a sphere and then generates a virtual perspective image for each object detection using a gnomonic projection, centered on the \ac{ROI}. Each virtual image is then processed through the network. }
    \label{fig:architecture}
    \vspace{-20pt}
\end{figure*}

Largely, works targeting deep learning methods applied to omni-directional or fisheye images resort to generating or collecting their own custom datasets, due to a lack of large scale annotated image datasets of this type. Many works simply project annotated perspective images to the distorted omni-directional or fisheye domain, as a simple way to generate a proxy dataset. Some recent work seeks to address this lack of annotated fisheye datasets~\cite{fu2019datasets, yogamani2019woodscape}. However, no such annotated dataset exists for fisheye images in the underwater domain. We address this problem by releasing our annotated UWHandles dataset, along with the method and annotation tool used to collect and process the images.

\section{METHOD}
\label{sec:method}
Special care must be taken in regressing 6D pose from full fisheye images, as there can be large distortions and ambiguity in the object viewpoint (Fig.~\ref{fig:viewpoints}). In the following sections, we outline how we use an intermediate spherical representation and the gnomonic projection to attain visually consistent pose annotations, followed by an overview of three different adaptions of the SilhoNet method\cite{billings2019silhonet} for 6D pose prediction from full fisheye images~(Fig.\ref{fig:architecture}).

\subsection{Spherical Mapping and Gnomonic Projection}

\begin{figure}
    \centering
    \includegraphics[width=1.0\linewidth]{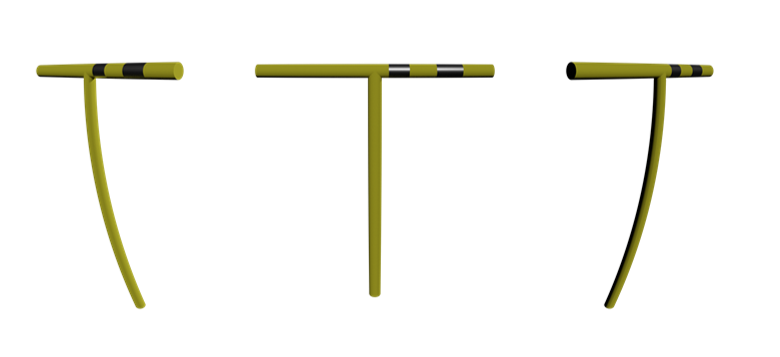}
    \vspace{-30pt}
    \caption{These objects have the same orientation relative to the rendered fisheye image frame but different translations, resulting in drastically different apparent orientations. Also, objects appear more distorted as they move from the image center.}
    \vspace{-15pt}
    \label{fig:viewpoints}
\end{figure}

While a class of different projection models exist for fisheye cameras~\cite{fisheye_model}, the model followed by the camera system used in this work, and the most common model in practice, is the equidistant projection
\vspace{-4pt}
\begin{equation}
    \vspace{-4pt}
    {\displaystyle R = f\theta}
\end{equation}
\noindent
where $\theta$ is the angle in radians from a point in the world to the optical axis, $f$ is the lens focal length, and $R$ is the radial position of the point projected on the imaging plane. A major challenge of fisheye images when regressing the object orientation is the large space of visual ambiguity. We define the global reference frame as coincident with the fisheye camera frame. As the angle between the object center in the world to the camera optical axis increases, there is increasing discrepancy between the object orientation relative to the global frame and the apparent orientation relative to a cropped \ac{ROI} (Fig.~\ref{fig:viewpoints}). We deal with this visual ambiguity by first mapping the fisheye image onto the unit sphere. The mapping between the pixel coordinates $(x,y)$ on the fisheye image with focal length $f$ and the polar coordinates $(\theta,\phi)$ on the unit sphere is given as
\vspace{-4pt}
\begin{equation}
    {\displaystyle r = \sqrt{x^2+y^2};\ \rho = r/f;\ z = \frac{r}{\tan{\rho}}}
\end{equation}
\noindent
\begin{equation}
    {\displaystyle \theta = \sin^{-1}{(\frac{y}{\sqrt{x^2+y^2+z^2}})};\ \phi = \tan^{-1}{\frac{x}{z}}}.
\end{equation}
\noindent
The inverse mapping can also be calculated by first converting the spherical coordinates to cartesian and then projecting onto the image plane with the fisheye model
\vspace{-4pt}
\begin{equation}
    {\displaystyle x_s = \cos{\theta}\sin{\phi};\ y_s = \sin{\theta};\ z_s = \cos{\theta}\cos{\phi}}
\end{equation}
\noindent
\begin{equation}
    {\displaystyle \rho = \cos^{-1}{(\frac{z_s}{\sqrt{x_s^2+y_s^2+z_s^2}})};\ r = f\rho}
\end{equation}
\noindent
\begin{equation}
    {\displaystyle x = \frac{x_sr}{\sqrt{x_s^2+y_s^2}};\ y = \frac{y_sr}{\sqrt{x_s^2+y_s^2}}.}
\end{equation}
\noindent
By mapping the fisheye image to a unit sphere centered on the global origin, we can define the apparent viewpoint of the object as the appearance of the object when projected onto a tangent plane centered on the vector extending from the sphere center to the center of the object. The projection from the sphere onto the tangent plane is known as a gnomonic projection and has a long history in mapping as well as recent application in omni-directional CNN methods~\cite{coors2018spherenet, zhao2018distortion}. Given a spherical mapping of an image and the tangent plane centered on the sphere at polar coordinates $(\theta_0,\phi_0)$, the gnomonic projection of the spherical point ($\theta$,$\phi$) onto the tangent plane is given as
\vspace{-4pt}
\begin{equation}
    {\displaystyle x = \frac{\cos{\theta}\sin{(\phi-\phi_0})}{\sin{\theta_0}\sin{\theta}+\cos{\theta_0}\cos{\theta}\cos{(\phi-\phi_0)}}}
\end{equation}
\noindent
\begin{equation}
    {\displaystyle y = \frac{\cos{\theta_0}\sin{\theta}-\sin{\theta_0}\cos{\theta}\cos{(\phi-\phi_0)}}{\sin{\theta_0}\sin{\theta}+\cos{\theta_0}\cos{\theta}\cos{(\phi-\phi_0)}}},
\end{equation}
\noindent
and an optimized inverse mapping from the tangent plane onto the sphere is given as
\vspace{-4pt}
\begin{equation}
    {\displaystyle \theta = \sin^{-1}{(\frac{\sin{\theta_0+y\cos{\theta_0}}}{\sqrt{1+x^2+y^2}}})}
\end{equation}
\noindent
\begin{equation}
    {\displaystyle \phi = \phi_0 +  \tan^{-1}{( \frac{x}{\cos{\theta_0}-y\sin{\theta_0}}})},
\end{equation}
\noindent
where $x$ and $y$ are the coordinates of the pixel on the tangent plane normalized by the virtual perspective camera focal length $f_p$~\cite{gnomonic_optimized, weisstein}. The gnomonic projection is core to our method of regressing the object 6D pose from \ac{ROI} proposals on the distorted fisheye image. The orientation of the object $R_p$ relative to a virtual perspective camera frame centered on the apparent viewpoint can be calculated as a rotation correction to the object orientation $R$ that is referenced to the global frame. The rotation correction matrix $R_{adj}$ can be constructed as follows. First, the polar coordinates $(\theta_0,\phi_0)$ of the intersection of the virtual camera optical axis with the sphere is calculated based on the 3D translation $(x,y,z)$ of the object relative to the global frame
\vspace{-4pt}
\begin{equation}
    {\displaystyle \theta_0 = \sin^{-1}{(\frac{y}{\sqrt{x^2+y^2+z^2}}})}
\end{equation}
\noindent
\begin{equation}
    {\displaystyle \phi_0 = \tan^{-1}{(\frac{x}{z}})}.
\end{equation}
\noindent
The rotation adjustment matrix is then constructed column-wise using the coordinates of the rotated virtual camera frame axes in the global reference frame
\vspace{-4pt}
\begin{equation}
    {\displaystyle X = [\cos{\phi_0},\ 0,\ -\sin{\phi_0}]}
\end{equation}
\noindent
\begin{equation}
    {\displaystyle Y = [-\sin{\theta_0}\sin{\phi_0},\ \cos{\theta_0},\ -\sin{\theta_0}\cos{\phi_0}]}
\end{equation}
\noindent
\begin{equation}
    {\displaystyle Z = [\cos{\theta_0}\sin{\phi_0},\ \sin{\theta_0},\ \cos{\theta_0}\cos{\phi_0}]}
\end{equation}
\noindent
\noindent
\begin{equation}
    {\displaystyle R_{adj} = [X;\ Y;\ Z]}
\end{equation}
\noindent
The orientation of the object relative to the virtual camera frame is then given as
\noindent
\begin{equation}
    {\displaystyle R_p = R_{adj}R}
\end{equation}
\noindent
The orientation branch of the network is trained to regress the apparent orientation $R_p$. The predicted true orientation $R$ can be recovered using the predicted object translation by constructing the inverse $R_{adj}$ matrix.

\subsection{SilhoNet Adaptation to Fisheye}

We compare three different variants of SilhoNet adapted for processing full fisheye images~(Fig.\ref{fig:architecture}). For all variants, the size of the predicted silhouettes was increased to 128x128, because the handle objects in the UWHandles dataset have very thin features. The translation prediction output was also modified to predict the normalized pixel offset of the object center relative to the \ac{ROI} directly without passing through a sigmoid function, and the predicted offsets were thresholded to lie within the \ac{ROI} bounds. Because the dataset does not include segmentation annotations, the occluded silhouette branch was removed from the network. The pose predictions of the handle objects were also reduced by their shape symmetries, as described in the SilhoNet paper~\cite{billings2019silhonet}. The annotated \ac{ROI}s were used as input to the network.

The first variant we consider as a baseline, which is essentially the vanilla SilhoNet architecture with the orientation branch output modified to regress the apparent orientation $R_{p}$, as described in the previous section. All variants of the network retain this prediction strategy. The second variant, which we refer to as "projective", processes the raw fisheye image through the feature extraction stage and then projects the features within the \ac{ROI} through a spherical mapping to the tangent plane centered on the \ac{ROI}, using the gnomonic projection. The projected features are then passed to the ROI-pooling stage. The motivating idea behind this projective strategy is that local features do not appear heavily distorted in fisheye images, but the spacial relationship of features across the \ac{ROI} can be significantly distorted. The local feature map is thus generated directly on the raw fisheye image and then the spacial relationship of these local features is corrected through the projection onto the tangent plane. This projection operation is implemented as a Tensorflow layer for efficient and simple integration into the original network. The third variant, which we refer to as "perspective", projects a region of the fisheye image to a virtual perspective image centered on the \ac{ROI} using the gnomonic projection. We chose the virtual image dimension to be 400x400 with a pixel relative focal length of 350. This virtual perspective image is processed through the feature extraction stage and then the \ac{ROI} is cropped from the center of the feature map and passed to the ROI-pooling stage. Essentially, this method generates a virtual perspective image for each detected object and processes each of these virtual images separately through the network. This methods corrects for the fisheye distortions through the entire network pipeline. However, the computation scales with the number of detected objects, as a separate virtual image is processed for each one.

As a further comparison point, we take each of the three variants described above and replace the silhouette prediction branch with a branch that directly regresses the quaternion orientation, rather than first predicting a silhouette and passing it to a second stage network for orientation prediction. This orientation branch has the same structure as the translation branch, but with the output size equal to 4x(\# classes). The predicted quaternion for the class of the detected object is extracted from the  output and normalized using an L2-norm. These methods which bypass the silhouette prediction to directly regress the orientation are referred to in the following sections by appending "\_direct" to the name of the associated variant: "baseline\_direct", "projective\_direct", and "perspective\_direct".

\subsection{Network Training}

The networks were trained with the same loss functions and dropout rates as in~\cite{billings2019silhonet} on Titan V GPUs. All networks were trained for 400,000 iterations on the training set except for the "perspective\_direct" method which was only trained for 356,000 iterations because of time constraints. Due to GPU memory limitations, the raw fisheye images of dimension 2,448x2,048 were downsampled by a factor of 3 for the baseline and projective variants and by a factor of 2 for the perspective variant. The baseline and projective variants were trained with a batch size of 2 and the perspective variant with a batch size of 3. As with the original SilhoNet method, the second stage network which regresses orientation from silhouettes was trained using only perfect rendered silhouettes.

\section{DATASET}
\label{sec:dataset}
We collected fisheye images of three different types of graspable handles, randomly arranged in different natural seafloor environments of the Costa Rican shelf break. Two of the handle types are actively used by Schmidt Ocean Institute and Woods Hole Oceanographic Institute to manipulate tools with Remotely Operated Vehicles (ROV) during underwater operations. AprilTag fiducials were randomly dispersed throughout the scene and on mount plates attached to the base of the handle objects in order to provide ground truth poses of the handles in the image sequences. The camera system was a FLIR BFLY-PGE-50S5C-C with a Fujinon FE185C086HA-1 fisheye lens centered in a dome housing that was mounted on the wrist of a Schilling Titan 4 hydraulic manipulator. This dataset is relevant to automating underwater manipulation tasks; if the pose of a handle attached to a known tool type can be accurately estimated, the handle can be autonomously grasped and manipulated to perform a desired manipulation task.

The dataset is composed of 25 training image sequences with a total of 18,329 images, 1 validation sequence with 910 images, and 2 testing sequences with 1,188 images.

\subsection{Data Collection}

At various locations on the seafloor, the ROV was set down and the handle objects were randomly dispersed throughout the reachable area of the manipulator. 4 metal tag plates with attached 4" AprilTag fiducial stickers were randomly scattered around the handle objects. Initially, we also attached 4" April Tag stickers to mount plates at the base of the handle objects, but through trial and error, we found the most robust detection results were obtained with multiple 2" stickers attached to the object mount plates rather than the single 4" sticker. The smaller tags on the objects enabled better detection of the tags at close range, and the large tags on the scattered metal plates provided good detection at greater distances. Full 5MP resolution images were recorded at 3Hz, with the manipulator moving around the objects in various motion paths to obtain a diverse set of viewpoints.

\subsection{Data Processing}

\begin{figure}
    \centering
    \includegraphics[width=1.0\linewidth]{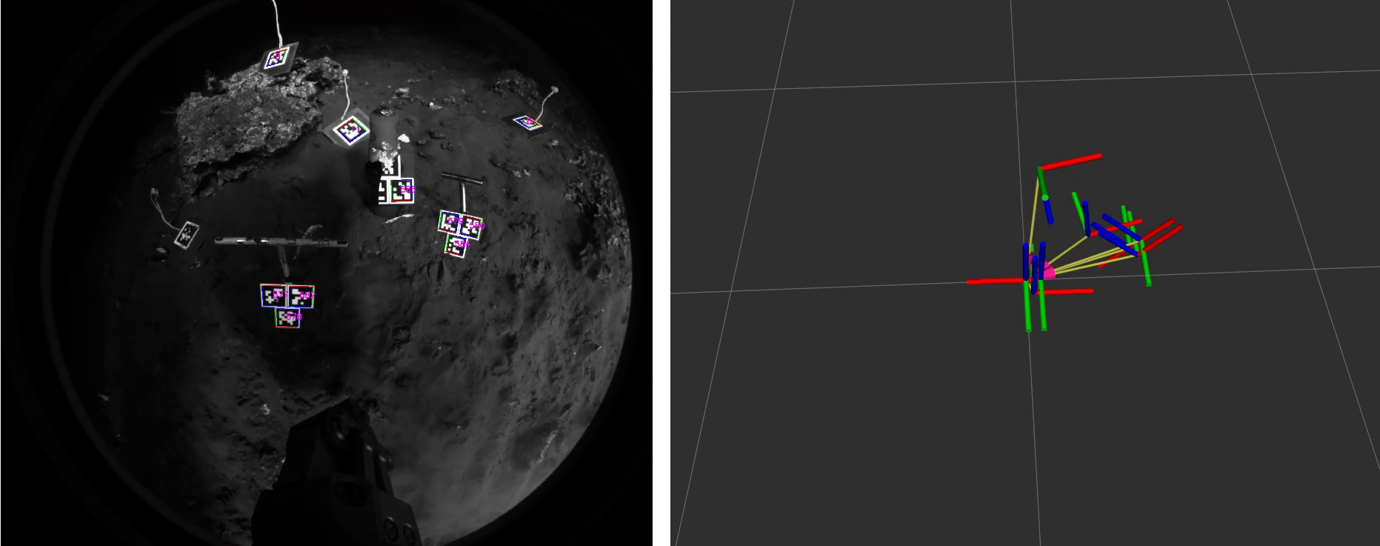}
    \vspace{-15pt}
    \caption{Tags are detected in the raw fisheye images and processed through TagSLAM~\cite{pfrommer2019tagslam} to get globally consistent camera poses for an image sequence.}
    \vspace{-20pt}
    \label{fig:tagslam}
\end{figure}

We used the ROS TagSLAM package~\cite{pfrommer2019tagslam} to process the image sequences and obtain globally consistent camera poses for each image in the sequence. In order to make use of the full fisheye view, tags were detected in the raw fisheye images (Fig.~\ref{fig:tagslam}), and then the detected tag poses were calculated using a pinhole equidistant distortion model calibrated with the Kalibr ROS package~\cite{maye2013self, furgale_maye_rehder}. The pinhole model was adequate for this camera system, because the effective usable field of view of the image was less than 180\degree.

\begin{figure}[h!]
    \vspace{-8pt}
    \centering
    \includegraphics[width=0.8\linewidth]{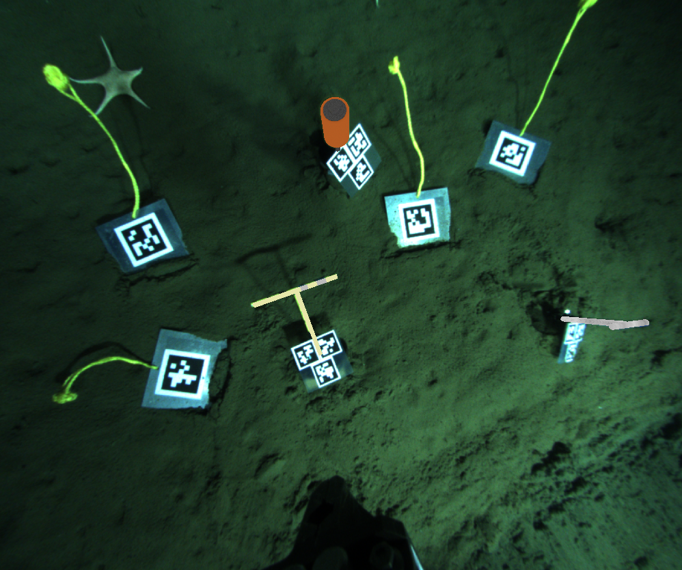}
    \vspace{-8pt}
    \caption{Sample from the UWHandles dataset showing the handle object models projected into the fisheye image using the 6D pose annotations. The fisheye image is rectified here for visualization.}
    \vspace{-15pt}
    \label{fig:poses}
\end{figure}

We created an OpenGL based annotation tool called VisPose, which takes in the image sequence with an associated camera pose file from the TagSLAM output. VisPose provides an interface to project models of the different objects into the image sequence, play through the sequence, and tweak the fit of the models to obtain accurate 6D pose annotations. Pose outliers can be filtered from the image sequence and then a COCO style annotation file exported, including 6D pose and 2D bounding box annotations, for the full image sequence (Fig.~\ref{fig:poses}).

\section{RESULTS}
\label{sec:results}
The following section presents the performance of the different SilhoNet adaptations on the UWHandles dataset. Table~\ref{tab:trans_thresh} and Table~\ref{tab:orient_thresh} show the percentage of translation and orientation predictions under different error thresholds, respectively. Table~\ref{tab:auc} shows the overall 6D pose prediction accuracy using the ADD-S metric from~\cite{xiang_posecnn:_2017}.

\begin{table}[h!]
  \scriptsize
  \centering
  \caption{Percentage of translation predictions under the threshold error, where a higher percentage under a lower threshold means better accuracy.}
  \vspace{-5pt}
    \begin{tabular}{lcccc}
    \toprule
    Method & \multicolumn{1}{P{3.5em}}{$<$ 5cm} & \multicolumn{1}{P{3.5em}}{$<$ 10cm} & \multicolumn{1}{P{3.5em}}{$<$ 20cm} & \multicolumn{1}{P{3.5em}}{$<$ 30cm} \\
    \midrule
    Baseline  & 69.88 & 90.81 & 98.48 & 99.92 \\
    Projective & 71.91 & 87.35 & 96.22 & 98.39 \\
    Perspective & 74.63 & 90.61 & 96.73 & 98.14 \\
    \midrule
    Baseline-Direct    & 47.55 & 72.04 & 94.56 & 99.21 \\
    Projective-Direct  & 46.71 & 73.56 & 93.54 & 98.36 \\
    Perspective-Direct & 57.69 & 81.29 & 96.11 & 99.07 \\
    \bottomrule
    \end{tabular}%
  \label{tab:trans_thresh}%
  \vspace{-5pt}
\end{table}%

\begin{table}[h!]
  \scriptsize
  \centering
  \caption{Percentage of orientation predictions under the threshold error, where a higher percentage under a lower threshold means better accuracy.}
  \vspace{-5pt}
    \begin{tabular}{lcccc}
    \toprule
    Method & \multicolumn{1}{P{3.5em}}{$<$ 5\degree} & \multicolumn{1}{P{3.5em}}{$<$ 10\degree} & \multicolumn{1}{P{3.5em}}{$<$ 20\degree} & \multicolumn{1}{P{3.5em}}{$<$ 30\degree} \\
    \midrule
    Baseline  & 12.75 & 34.77 & 62.78 & 75.31 \\
    Projective & 11.26 & 35.33 & 63.03 & 74.89 \\
    Perspective & 16.05 & 39.08 & 66.08 & 77.28 \\
    \midrule
    Baseline-Direct    & 22.58 & 45.58 & 69.00 & 81.31 \\
    Projective-Direct  & 28.91 & 50.45 & 69.48 & 83.19 \\
    Perspective-Direct & 29.81 & 55.01 & 74.21 & 85.02 \\
    \bottomrule
    \end{tabular}%
  \label{tab:orient_thresh}%
  \vspace{-5pt}
\end{table}%

\begin{table}[h!]
  \scriptsize
  \centering
  \caption{Area under accuracy-threshold curve for 6D pose evaluation using ADD-S metric from \cite{xiang_posecnn:_2017}, where a higher area means better accuracy. Proj. is short for Projective and Persp. is short for Perspective}
  \vspace{-5pt}
    \begin{tabular}{lccc|ccc}
    \toprule
    Handle Type & \multicolumn{1}{P{3em}}{Baseline} & \multicolumn{1}{P{.5em}}{Proj.} & \multicolumn{1}{P{.5em}}{Persp.} & \multicolumn{1}{|P{2.8em}}{Baseline Direct} & \multicolumn{1}{P{2.5em}}{Proj. Direct} & \multicolumn{1}{P{2.5em}}{Persp. Direct} \\
    \midrule
    umichhandle & 72.71 & 69.53 & 78.81 & 61.70 & 61.79 & 64.46 \\
    soihandle   & 71.48 & 79.98 & 75.11 & 47.51 & 53.33 & 60.77 \\
    whoihandle  & 61.82 & 57.34 & 61.95 & 48.54 & 47.39 & 59.90 \\
    \midrule
    ALL & 68.65 & 68.92 & 71.94 & 52.57 & 54.15 & 61.69 \\
    \bottomrule
    \end{tabular}%
  \label{tab:auc}%
  \vspace{-20pt}
\end{table}%

For translation prediction errors under 5cm, which is a common measure of interest for pose estimation methods, the perspective variant shows significant performance improvement over the baseline method, while the projective method shows some improvement. All of the direct variants that remove the intermediate silhouette prediction branch show a drastic drop in translation prediction accuracy, indicating that even though the silhouettes are not directly used in the translation prediction, they enhance the networks ability to learn accurate feature scaling. The perspective-direct variant still shows significant improvement over the baseline-direct method, indicating that compensating for distortions in the fisheye image rather than directly predicting from the raw image is important for accurate pose predictions.

For orientation prediction errors under 5\degree, the perspective variant also shows significant performance improvement over the basline method, while the projective variant does not perform as well as the baseline. In contrast to the translation predictions, all of the direct methods improve on the orientation prediction accuracy by approximately a factor of two across all variants, while the perspective-direct method still outperforms the baseline-direct method by a large margin.

The ADD-S results also show a strong improvement in performance for the perspective variant against the baseline, both with and without the silhouette predictions, while the projective and baseline methods perform similarly. Because the ADD-S metric is generally most sensitive to translation errors, the results show stronger performance for the methods that retain the intermediate silhouette prediction over the direct methods. However, taking into account the separate orientation and translation results, better overall performance on this dataset might be achieved by a method which directly predicts the orientation but retains a silhouette prediction branch during training to boost the translation accuracy. This is a method we plan to explore in the future. Overall, the results indicate that accounting for fisheye distortions before feature extraction, as the perspective method does, gives the best performance.

\section{Conclusion}
\label{sec:conclusion}
In this paper, we presented a framework for adapting a \ac{ROI}-based 6D object pose estimation method to work on full fisheye images. We demonstrated the adaptation of the SilhoNet~\cite{billings2019silhonet} method on a new dataset of annotated fisheye images, called UWHandles, collected in natural underwater seafloor environments. The objects in the dataset are visually challenging handles, used in ROV operations to manipulate tools. The testing results on this dataset show that directly accounting for the fisheye distortions in the network before feature extraction is important for improving pose prediction accuracy, where the best performance was obtained with a method that generates a virtual perspective image centered on each \ac{ROI} detection, and processes these virtual undistorted images separately through the network. The results also show that the intermediate silhouette predictions of the SilhoNet method are important for the network to learn feature scaling to accurately predict translation. However, for this dataset, directly regressing the orientation rather than predicting from an intermediate silhouette achieves the greatest orientation accuracy. These observations motivate future investigation into a method which directly regresses the orientation, but retains the silhouette prediction branch during training to supervise the learning of feature scaling.

\section{Acknowledgments}
\label{sec:acknowledgments}
This work was supported by the NASA PSTAR grant NNX16AL08G.

\renewcommand{\bibfont}{\normalfont\small}
\printbibliography

\end{document}